\documentclass[a4paper]{article}

\usepackage{INTERSPEECH_v2}
\usepackage{cite}

\title{Direct Acoustics-to-Word Models\\for English Conversational Speech Recognition}
\name{Kartik Audhkhasi, Bhuvana Ramabhadran, George Saon, Michael Picheny, David Nahamoo\thanks{The authors  thank Yajie Miao and Florian Metze of Carnegie Mellon University, and Stanley Chen of IBM for helpful discussions.}}
\address{IBM T. J. Watson Research Center, Yorktown Heights, NY}
\email{\{kaudhkha,bhuvana,saon,picheny,nahamoo\}@us.ibm.com}

\begin{document}

\maketitle
\begin{abstract}
Recent work on end-to-end automatic speech recognition (ASR) has shown that the connectionist temporal classification (CTC) loss can be used to convert acoustics to phone or character sequences. Such systems are used with a dictionary and separately-trained Language Model (LM) to produce word sequences. However, they are not truly end-to-end in the sense of mapping acoustics directly to words without an intermediate phone representation. In this paper, we present the first results employing direct acoustics-to-word CTC models on two well-known public benchmark tasks: Switchboard and CallHome. These models do not require an LM or even a decoder at run-time and hence recognize speech with minimal complexity. However, due to the large number of word output units, CTC word models require orders of magnitude more data to train reliably compared to traditional systems. We present some techniques to mitigate this issue. Our CTC word model achieves a word error rate of 13.0\%/18.8\% on the Hub5-2000 Switchboard/CallHome test sets without any LM or decoder compared with 9.6\%/16.0\% for phone-based CTC with a 4-gram LM. We also present rescoring results on CTC word model lattices to quantify the performance benefits of a LM, and contrast the performance of word and phone CTC models.

\end{abstract}

\section{Introduction}\label{sec:intro} 
Feed-forward, recurrent, and convolutional deep neural networks (DNNs) have significantly improved the state-of-the-art in acoustic models (AMs)~\cite{hinton2012deep,saon2017english,xiong2016microsoft}. Advanced neural network language models (LMs)~\cite{mikolov2011empirical, chelba2013one} and exponential LMs~\cite{chen2009shrinking} have also significantly out-performed count-based N-gram LMs. Despite these advances, building a state-of-the-art automatic speech recognition (ASR) system is still a cumbersome multi-step exercise. This is fundamentally linked to the mathematical framework under which current ASR systems operate. ASR is typically formulated as the following maximum a-posteriori (MAP) optimization problem of finding the best word sequence given the acoustics~\cite{jelinek1997statistical}:
\begin{align}\label{eq:asr}
	\mathbf{w}_\text{MAP} &= \arg\max_{\mathbf{w}} P_\text{ASR}(\mathbf{w}|\mathbf{A};\Theta_\text{ASR}) \;.
\end{align}
Here $\mathbf{A}$ is a $T \times D$ matrix of $D$-dimensional acoustic feature vectors over $T$ time steps and $\mathbf{w}$ is the sequence of words. 
 Bayes' theorem splits this model into two:
\begin{align}\label{eq:bayes}
	P_\text{ASR}(\mathbf{w}|\mathbf{A};\Theta_\text{ASR}) &\propto P_\text{AM}(\mathbf{A}|\mathbf{w};\Theta_\text{AM}) P_\text{LM}(\mathbf{w};\Theta_\text{LM}) \;.
\end{align}
Here $P_\text{AM}(\mathbf{A}|\mathbf{w};\Theta_\text{AM})$ gives the AM probability of the acoustics $\mathbf{A}$ given the words $\mathbf{w}$, and $P_\text{LM}(\mathbf{w};\Theta_\text{LM})$ gives the LM probability of the word sequence $\mathbf{w}$. The ASR training process thus splits into two disjoint learning problems, one each for the AM and the LM. It makes decoding a convoluted process of the fusion of AM and LM log-likelihoods for many candidate word sequences before picking the most likely word sequence. Training the popular "hybrid" DNN-hidden Markov Model (HMM) AMs also involves multiple steps. 

Recent research in neural networks has led to "end-to-end" learning which aims to simplify the above process. This includes the connectionist temporal classification (CTC) loss~\cite{graves2006connectionist} and the encoder-decoder framework~\cite{bahdanau2014neural,bahdanau2016end}. In particular, the CTC loss enables training AMs by mapping acoustics to phone or character sequences without any frame-level alignment~\cite{maas2015lexicon, graves2014towards, hannun2014deep, miao2015eesen, miao2016empirical, sak2015fast, zweig2016advances}.  
However, phone/character-based CTC AMs are not truly end-to-end acoustics-to-word because they require a dictionary and an externally-trained LM during decoding to perform well.

Direct acoustics-to-word CTC models are a natural step towards true end-to-end ASR. The work by Sak et al.~\cite{sak2015fast}, and the more recent work by Soltau, Liao, and Sak~\cite{soltau2016neural} has presented direct acoustics-to-word CTC models. The latter trained word CTC models on around 125,000 hours of speech from a YouTube data set and a vocabulary of 100,000 words, and achieved results close to the state-of-the-art. 
This raises many interesting research questions: How well do word CTC models perform on standard public benchmarks that have advanced the state-of-the-art in ASR over the last few decades? Do they only work in a brute-force data-intensive setting? How do word CTC models scale across languages and in low-resource conditions?

To help address these questions, we present the first results using direct acoustics-to-word CTC models on two well-known public benchmark data sets - Switchboard and CallHome. CTC word models require orders of magnitude more data to train reliably compared to phone/character-based models. We present two techniques to improve the training of these models on 300 hours of Switchboard and 2000 hours of Switchboard+Fisher training sets. These include incorporating
\begin{enumerate}
\item Phone CTC models and hierarchical CTC~\cite{fernandez2007sequence}.
\item GloVe word embeddings~\cite{pennington2014glove} to capture LM-like word co-occurrence information. 
\end{enumerate}
We find that both these techniques improve training convergence and performance of the word CTC models compared to random initialization. Our CTC word model achieves a word error rate of 13.0\%/18.8\% on the Hub5-2000 Switchboard/CallHome test sets without any LM or decoder compared with 9.6\%/16.0\% for phone-based CTC with a 4-gram LM. 
 
The next section presents our phone CTC system, while Section~\ref{sec:word_ctc} describes our word CTC system, various strategies we adopted to improve its performance, and an error-based comparison of the word and phone CTC systems. We conclude the paper in Section~\ref{sec:concl} with directions for future work.

\section{Baseline Phone CTC System}\label{sec:phone_ctc}
Before discussing our phone CTC system, we first present a quick overview of the CTC loss function~\cite{graves2006connectionist} for completeness.

\subsection{CTC Loss}\label{sec:ctc_loss}
Let $\mathbf{y}$ denote the length-$L$ target symbol sequence consisting of phones, characters, or words. Let $\mathbf{A}$ denote the $T \times D$ matrix of $D$-dimensional acoustic feature vectors over $T$ time steps. The conventional cross-entropy loss requires $L$ to be equal to $T$. A common approach to solve this problem is by force-aligning $\mathbf{y}$ with $\mathbf{A}$, which yields a mapping from any time $t \in \{1,\ldots,T\}$ to a symbol $y_l \in \mathbf{y}$. Instead, CTC allows an extra "blank" symbol $\phi$ that expands the length-$L$ sequence $\mathbf{y}$ to a set of length-$T$ sequences $\Omega(\mathbf{y})$, where each sequence $\tilde{\mathbf{y}} \in \Omega(\mathbf{y})$ reduces to $\mathbf{y}$ after the following two operations in sequence:
	\begin{enumerate}
		\item Removal of all repeating symbols.
		\item Removal of the blank symbol $\phi$.
	\end{enumerate}
The set $\Omega(\mathbf{y})$ is often called the set of CTC "paths" corresponding to the symbol sequence $\mathbf{y}$. The negative CTC loss is then
\begin{align}\label{eq:ctc_loss}
	P(\mathbf{y}|\mathbf{A}) &= \sum_{\tilde{\mathbf{y}}\in\Omega(\mathbf{y})} P(\tilde{\mathbf{y}}|\mathbf{A}) = \sum_{\tilde{\mathbf{y}}\in\Omega(\mathbf{y})} \prod_{t=1}^T P(\tilde{y}_t|\mathbf{a}_t) \;.
\end{align} 
Dynamic programming efficiently computes the above function using the forward-backward recursion.
It is easy to see that CTC implicitly constructs a left-to-right HMM with $2L+1$ states for a $L$-length sequence $\mathbf{y}$ by interpolating the symbol states with blank states that can be optionally skipped. A neural network predicts the probability of occupying one of the $2L+1$ HMM states at each of the $T$ time steps. The forward-backward algorithm on this trellis computes the CTC loss (\ref{eq:ctc_loss}) and its gradients, which are then back-propagated through the neural network~\cite{graves2006connectionist}.

\subsection{Training and Testing Data Sets}
We used 262 hours of segmented speech from the standard 300-hour Switchboard-1 audio with transcripts provided by Mississippi State University for training the 300-hour systems. We added 1698 hours of audio from the Fisher data collection and 15 hours from the CallHome audio to build the (approximately) 2000-hour systems. We built two LMs for decoding the phone CTC models and rescoring the word CTC models. The "small" 4-gram LM used 24M words from the 2000-hour audio training data with a vocabulary size of 30k words, while the "large" 4-gram LM used a vocabulary of 85k words with an additional 560M words from several public text data sets from LDC~\cite{saon2017english}.

\subsection{Phone CTC Model}
We trained our phone CTC system over $44$ phones from the Switchboard pronunciation lexicon~\cite{saon2017english} plus the blank symbol. We extracted 40-dimensional logMel filterbank energies over 25 ms frames every 10 ms from the input speech signal, stacked two successive frames, and dropped every alternate frame resulting in $80$-dimensional logMel features at half the rate of the original 40-dimensional features. This "stacking+decimation" operation ~\cite{sak2015fast} provided significant speed-up in training because the sequence length reduces by half with no loss in performance.
We also used 100-dimensional i-vectors for each speaker~\cite{dehak2011front} and appended them to each feature, resulting in 180-dimensional feature vectors.

We used the Torch toolkit~\cite{torch} with the cuDNN v5~\cite{chetlur2014cudnn} backend to build bidirectional long short-term memory (BLSTM) networks with 5 layers and 320 neurons each in the forward and backward layers. This BLSTM feeds into a fully-connected linear layer of size 640$\times$45 followed by the softmax activation function. We used the Torch binding of the Warp-CTC tool~\cite{warp_ctc} for computing the CTC loss and its gradients. We prepared the training data by sorting the utterances in decreasing order of number of frames~\cite{miao2015eesen}, and dividing them into batches of 48 utterance each. We used stochastic gradient descent (SGD) with a learning rate of $5\times10^{-4}$, and cut the learning rate by half whenever the heldout loss did not decrease by more than $10\%$. All parameters of the neural network assumed uniformly random initial values in $(-0.01,0.01)$. We also clipped the gradients to lie in $(-10,10)$ to stabilize training. All our models converged in 15-20 epochs.

We constructed our CTC decoding graph similar to the one used in~\cite{miao2015eesen}. Table~\ref{tab:phone_ctc} shows the word error rate (WER) of our phone CTC system on the Hub5-2000 Switchboard and CallHome test sets. We see that using the big LM gives a consistent gain of around 0.5\% absolute in WER.  For reference, our best 6-layer 2000-hour hybrid BLSTM using FMLLR+i-vector features with 32,000 context-dependent states with standard cross-entropy+sequence training has a WER of 7.7\%/14.0\% on Switchboard/CallHome~\cite{saon2017english} when decoding with the big LM. However, decoding with the phone CTC model is almost 2x faster and 8x less memory-intensive than the hybrid BLSTM model.

\begin{table}[h]
\caption{This table shows the WERs of our phone CTC models on the Switchboard (SWB) and CallHome (CH) Hub5-2000 test sets for both the "small" and "big" N-gram LMs.}
\begin{center}
\begin{tabular}{|c|c|c|c|} \hline
\emph{Hours} & \emph{LM} & \emph{SWB} & \emph{CH} \\ \hline\hline
300 &  Small & 14.5 & 25.1 \\ \hline
300 & Big & {\bf 13.9} & {\bf 24.7} \\ \hline
2000 & Small & 10.2 & 16.5 \\ \hline
2000 & Big & {\bf 9.6} & {\bf 16.0} \\ \hline
\end{tabular}
\end{center}
\label{tab:phone_ctc}
\end{table}

\section{Word CTC Model}\label{sec:word_ctc}
Word CTC models require orders of magnitude more data to train reliably compared to conventional ASR systems. To mitigate this huge need for data, we restricted the vocabulary to contain words with at least 5 acoustic examples in the training data. This resulted in a vocabulary of approximately 10,000 words for the 300-hour system and approximately 25,000 words for the 2000-hour system. We created a special token to denote all out-of-vocabulary (OOV) words. The OOV rate for the 300-hour training data was 1.5\% with the 10,000 word vocabulary, while the corresponding OOV rate for the 2000-hour training data was 0.5\% for the 25,000 word vocabulary. Decoding involved a simple forward pass of the acoustic sequence through the network followed by picking the highest-scoring word at each time step and repetition/blank removal. We mapped the OOV token to the silence symbol for scoring.

Similar to the phone CTC model, we setup a 5-layer BLSTM for the word CTC model with 320 forward and backward hidden neurons in each layer. The final dense layer mapped 640 dimensions to the vocabulary size+1 (for the blank symbol). Our first training attempt used uniformly-random weight initialization over $(-0.01,0.01)$ similar to the phone CTC model. However, the training failed to converge as shown in Figure~\ref{fig:word_ctc_loss}, despite tuning the learning rate and the amount of gradient clipping. We hypothesized that training the BLSTM to learn a direct mapping from acoustics to words from flat-start is a difficult proposition, especially for small data sets. We next discuss some techniques we experimented with on the 300-hour set to mitigate this issue. 

\subsection{BLSTM Initialization with Phone CTC Model and Hierarchical CTC}
As a first step, we initialized the word CTC BLSTM with the phone CTC BLSTM, with the intuition that the ability to detect sub-word units provides a good starting point for detecting words. We initialized the final dense layer randomly as before. Figure~\ref{fig:word_ctc_loss} shows the training and validation CTC loss using random initialization and with initialization using the phone BLSTM network. The word CTC model fails to converge with random initialization (red curves) but converges nicely when using the phone BLSTM model as a starting point (green curves).

We also experimented with hierarchical CTC~\cite{fernandez2007sequence,rao2017multi} where the bottom 4 BLSTM layers were initialized with a pre-trained phone CTC model and the top BLSTM layer is randomly initialized. Training then proceeds in multi-task fashion with a weight of $\alpha$ to the phone CTC loss computed by branching-off from the 4th BLSTM layer and $1-\alpha$ to the final word CTC loss. After several experiments with different $\alpha$, we observed that while hierarchical CTC was able to out-perform full random initialization, it was always worse compared with simply initializing the entire BLSTM network with the phone CTC BLSTM and training with only the word CTC loss.

\subsection{Dense Layer Initialization with Word Embeddings}
Encouraged by the impact of initializing the word BLSTM with the phone BLSTM, we wanted to explore better ways of initializing the final dense layer as well. The weights of the final layer are a set of 10,000 640-dimensional vectors, one for each word in the vocabulary. At each time step, the dense layer computes a dot product between the hidden representation and the weight vector corresponding to each word, resulting in a scalar score for each word. Words co-occurring frequently in natural text are likely to have high scores closer to each other in time. We can thus think of each 640-dimensional weight vector as a word embedding that captures word co-occurrence information. 

The above intuition led us to use GloVe embeddings~\cite{pennington2014glove} for initializing the final dense layer, similar to the technique proposed in~\cite{audhkhasi2016semantic} for augmenting neural network LMs with externally-trained GloVe embeddings. GloVe is similar in spirit to word2vec~\cite{mikolov2013efficient} because it also tries to capture word co-occurrence information. However, Glove achieves this through a bilinear approximation 
of the word co-occurrence matrix. 
%
%

\begin{figure}[htbp]
\begin{center}
\includegraphics[width=0.40\textwidth]{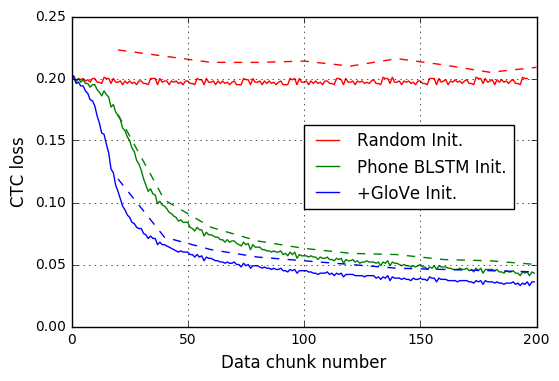}
\caption{This figure shows the word CTC model training (solid) and heldout (dashed) losses for the first 10 epochs over the 300-hour training set using different initialization schemes. Each data chunk on the x-axis corresponds to roughly 15 hours.}
\label{fig:word_ctc_loss}
\end{center}
\end{figure}

We trained two sets of 640-dimensional GloVe embeddings: one on the 24M word corpus used for training the small LM, and another on the 560M word corpus used for the big LM. We assigned random vectors for the blank symbol and OOV token. We normalized each word vector to have a unit L2 norm and scaled it by 0.1. Figure~\ref{fig:word_ctc_loss} shows that the GloVe initialization of the final dense layer by the average of the small and big LM embeddings yields significant improvement in training and heldout CTC loss. Table~\ref{tab:word_ctc_init} shows a 3.6\%/3.7\% absolute improvement in WER for Switchboard/CallHome by using GloVe initialization over the phone BLSTM initialization. We also note that GloVe embeddings trained on the bigger 24M+540M word data set yield around 1\% improvement in WER over the embeddings trained only on the 24M word set.

\begin{table}[ht]
\caption{This table shows the Switchboard and CallHome WERs for different initialization schemes for the 300 hour CTC word models. The randomly-initialized model failed to converge.}
\begin{center}
\begin{tabular}{|c|c|c|} \hline
\emph{Init. Method} & \emph{SWB} & \emph{CH} \\ \hline\hline
Random Init. & - & - \\ \hline
Phone BLSTM Init. & 24.4 & 34.1 \\ \hline
+ GloVe Init. (24M) & 21.7 & 31.5 \\ \hline
+ GloVe Init. (560M) & {\bf 20.8} & {\bf 30.4} \\ \hline
\end{tabular}
\end{center}
\label{tab:word_ctc_init}
\end{table}

\begin{figure*}[htbp]
\begin{center}
\includegraphics[width=0.9\textwidth]{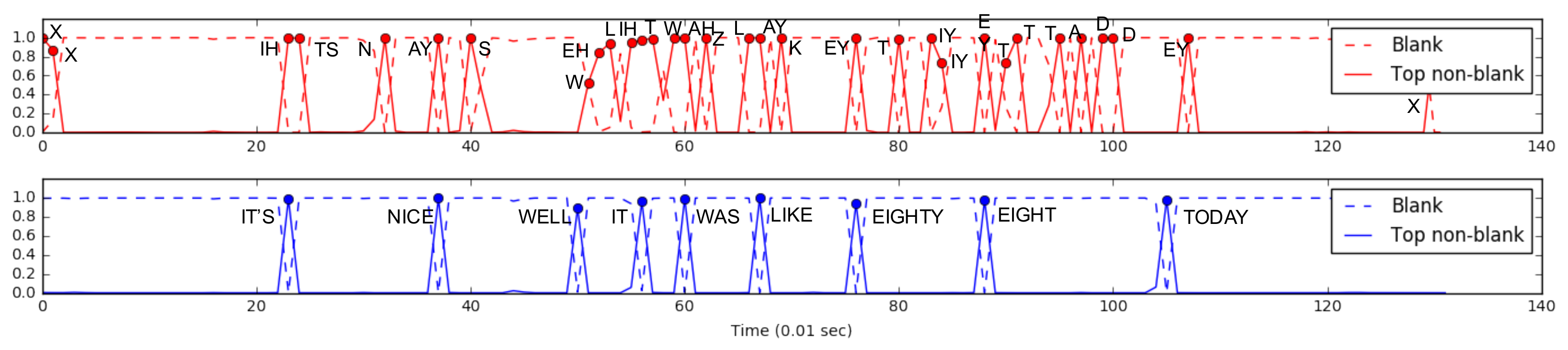}
\caption{This figure shows the 2000-hour CTC phone (red, top) and CTC word (blue, bottom) model posteriors for an utterance from the Switchboard test set, along with the phones and words in the respective 1-best paths. "X" denotes the silence phone.}
\label{fig:ctc_post_plot}
\end{center}
\end{figure*}

\subsection{Final 2000-Hour Word CTC Model}
We initialized the 2000-hour word CTC model with the BLSTM from the trained 2000-hour phone CTC model and GloVe embeddings trained over the 24M+560M word text corpus. We used the same input acoustic featurs as before with frame stacking+decimation. The output vocabulary was 25,000 words with a training set coverage of approximately 99.5\%.

\begin{table}[ht]
\caption{This table shows the Switchboard/CallHome WERs for our 2000-hour word and phone CTC models.}
\begin{center}
\begin{tabular}{|c|c|c|c|} \hline
\emph{AM} & \emph{LM} & \emph{SWB} & \emph{CH} \\ \hline\hline
Word CTC & - & {\bf 13.0} & {\bf 18.8} \\ \hline
Word CTC & Big (rescoring) & {\bf 12.5} & {\bf 18.0} \\ \hline\hline
Phone CTC & Big &  9.6 & 16.0 \\ \hline
\end{tabular}
\end{center}
\label{tab:word_ctc_wer}
\end{table}

Table~\ref{tab:word_ctc_wer} compares the performance of our 2000-hour word CTC model with the big LM results from the 2000-hour phone CTC model. We observe that our 2000-hour word CTC system achieves a WER of 13.0\%/18.8\% on Switchboard/CallHome without using any LM, while the best phone CTC system using the big LM achieves 9.6\%/16.0\%. This result is encouraging since recognizing speech using the word CTC model uses no LM or decoder, and hence takes significantly less time and memory compared with the phone CTC model. It is interesting to note that the gains in WER upon increasing the amount of training audio from 300 to 2000 hours is bigger for the word CTC model (7.8\%/11.6\%) compared to the phone CTC model (4.3\%/8.7\%). This is due to word targets receiving more training data, and also due to the 1\% reduction in OOV rate upon increasing the number of word targets from 10,000 to 25,000.


\subsection{Word CTC LM Rescoring}
To gauge the impact of using a LM, we rescored the output of the word CTC model with the big LM. The word CTC model posteriors are very spiky, as shown in Figure~\ref{fig:ctc_post_plot}, with the top-scoring word or blank symbol getting most of the probability mass at each frame and the blank symbol dominating an overwhelming majority of frames. We picked the time locations with words as the top hypotheses, and took the next best $K-1$ words at these frames. This gave us a consensus network or sausage-like lattice containing as many nodes as the number of words in the 1-best hypotheses. We also subtracted the word log-priors computed over the training set from the acoustic log-posterior score at each arc to give acoustic log-likelihood.

Table~\ref{tab:word_ctc_wer} shows the WERs after rescoring the word CTC lattices with the big LM for $K = 10$. Higher values of $K$ did not give more gains. We observe a 0.5\%/0.8\% absolute improvement in WER, which is similar to what we got upon using the big LM to decode phone CTC models in Table~\ref{tab:phone_ctc}. However, the remaining 2.9\%/2\% gap between the rescored word model and the phone CTC model indicates that some work needs to be done on improving the word CTC model itself. The oracle WERs of the Switchboard and CallHome word CTC lattices was 6.8\% and 11.4\%, which indicates that the CTC word system can potentially achieve a significantly lower WER.

\subsection{Comparison of Word and Phone CTC System Errors}\label{sec:comp}
As an initial comparison between the word and phone CTC models, we wanted to see if there is any big discrepancy between the two based on the type of errors committed. Table~\ref{tab:error_rates} shows the substitution, deletion, and insertion rates for the 2000-hour phone CTC and word CTC models (with and without big LM rescoring). We observe that word CTC lags behind the phone CTC model most often in the deletion rate. With the exception of the unrescored word CTC model, the substitution and insertion rates for the phone and word CTC models differ by less than a percent. 

\begin{table}[ht]
\caption{This table shows substitution, deletion, and insertion rates of the 2000-hour word and phone CTC systems on the Switchboard and CallHome test sets. Numbers in parentheses show absolute differences with respect to the phone CTC model.}
\begin{center}
\begin{tabular}{|c|c|c|c|} \hline
\emph{Model} & \emph{SWB Sub.} & \emph{SWB Del.} & \emph{SWB Ins.} \\ \hline\hline
Phone CTC & 5.8 & 2.7 & 1.1 \\ \hline
Word CTC & {\bf 7.4 (+1.6)} & 4.1(+1.4) & 1.6 (+0.5) \\ \hline
+LM resc. & 6.4 (+0.6) & {\bf 4.6 (+1.9)} & 1.5 (+0.4) \\ \hline\hline
\emph{Model} & \emph{CH Sub.} & \emph{CH Del.} & \emph{CH Ins.} \\ \hline\hline
Phone CTC & 9.6 & 4.7 & 1.7 \\ \hline
Word CTC & 10.3 (+0.7) & {\bf 6.4 (+1.7)} & 2.1(+0.4) \\ \hline
+LM resc. & 10.0 (+0.4) & {\bf 6.1 (+1.4)} & 1.9 (+0.2) \\ \hline
\end{tabular}
\end{center}
\label{tab:error_rates}
\end{table}

We also did a per-utterance comparison of the errors committed by the 2000-hour word and phone CTC systems. For 61.9\%/61.1\% of utterances from the Switchboard/CallHome test sets, the word and phone CTC systems commit the same number of errors. They commit identical errors in 42\%/41\% of the utterances. The word CTC system makes at most (less than or equal to) the same number of errors as the phone CTC system in 72.1\%/74.9\% of the utterances. Hence for a sizable majority of test utterances, the word CTC system makes equal or fewer errors than the phone CTC system, and is much faster to decode with. 


\section{Conclusion and Future Work}\label{sec:concl}
This paper presented the first results using direct acoustics-to-word CTC models on the well-benchmarked Switchboard and CallHome data sets. We presented two techniques for initializing the word CTC models and improving their performance - (1) using the phone CTC BLSTM and hierarchical CTC, and (2) using GloVe word embeddings for initializing the final dense layer. Our 2000-hour word CTC system achieved a WER of 13.0\%/18.8\% on the Switchboard/CallHome data sets, compared to 9.6\%/16.0\% for the phone CTC system. We also showed the impact of lattice rescoring on the word CTC model. We also observed that the word CTC system is at least as good as the phone CTC system in around 70\% of the utterances for both the test sets, while being significantly faster to decode with due to lack of a LM. 

Future work will focus on closing the gap between the word and phone CTC systems without scaling the amount of training data by several orders of magnitude, exploring techniques for speeding-up the training of word CTC models further, and evaluating word CTC models for other languages and tasks, especially low-resource ones.

\bibliographystyle{IEEEtran}

\bibliography{ibm_ctc_interspeech2017}

\end{document}